\documentclass[]{spie}  %>>> use for US letter paper
\usepackage[]{graphicx}

\usepackage{amsmath}
\usepackage{amssymb}
\usepackage{color,soul} % highlighting
\usepackage[hidelinks,bookmarksnumbered,pdfhighlight=/P]{hyperref} %% nicest :-)

\usepackage{float} % stop figure from floating with [htb!] (package {here} is obsolete

\usepackage{booktabs} % To thicken table lines
\usepackage{rotating}
\usepackage{multirow}

\def \Npatients {82 }
\def \Ntrain {60 }
\def \Nvalid {2 }
\def \Ntest {20 }

\title{Deep convolutional networks for pancreas segmentation in CT imaging} 

\author{Holger R. Roth, Amal Farag, Le Lu, Evrim B. Turkbey, and Ronald M. Summers
\skiplinehalf
Imaging Biomarkers and Computer-Aided Diagnosis Laboratory\\ 
Radiology and Imaging Sciences\\
National Institutes of Health Clinical Center\\
Bethesda, MD 20892-1182, USA.
}

\authorinfo{Further author information: Send correspondence to Holger Roth (\url{holger.roth@nih.gov}) or Ronald Summers (\url{rms@nih.gov})}
\begin{document} 
  \maketitle 

%%%%%%%%%%%%%%%%%%%%%%%%%%%%%%%%%%%%%%%%%%%%%%%%%%%%%%%%%%%%% 
\begin{abstract}
Automatic organ segmentation is an important prerequisite for many computer-aided diagnosis systems. The high anatomical variability of organs in the abdomen, such as the pancreas, prevents many segmentation methods from achieving high accuracies when compared to state-of-the-art segmentation of organs like the liver, heart or kidneys. Recently, the availability of large annotated training sets and the accessibility of affordable parallel computing resources via GPUs have made it feasible for ``deep learning'' methods such as convolutional networks (ConvNets) to succeed in image classification tasks. These methods have the advantage that used classification features are trained directly from the imaging data.

We present a fully-automated bottom-up method for pancreas segmentation in computed tomography (CT) images of the abdomen. The method is based on hierarchical coarse-to-fine classification of local image regions (superpixels). Superpixels are extracted from the abdominal region using Simple Linear Iterative Clustering (SLIC). An initial probability response map is generated, using patch-level confidences and a two-level cascade of random forest classifiers, from which superpixel regions with probabilities larger 0.5 are retained. These retained superpixels serve as a highly sensitive initial input of the pancreas and its surroundings to a ConvNet that samples a bounding box around each superpixel at different scales (and random non-rigid deformations at training time) in order to assign a more distinct probability of each superpixel region being pancreas or not. 

We evaluate our method on CT images of \Npatients patients (\Ntrain for training, \Nvalid for validation, and \Ntest for testing). Using ConvNets we achieve average Dice scores of 68\% $\pm$ 10\% (range, 43-80\%) in testing. This shows promise for accurate pancreas segmentation, using a deep learning approach and compares favorably to state-of-the-art methods.

\end{abstract}
%%%%%%%%%%%%%%%%%%%%%%%%%%%%%%%%%%%%%%%%%%%%%%%%%%%%%%%%%%%%%
%%%%%%%%%%%%%%%%%%%%%%%%%%%%%%%%%%%%%%%%%%%%%%%%%%%%%%%%%%%%%
\section{Introduction}
\begin{figure}[htb!]%[htb!]
\centering	\resizebox{0.3\textwidth}{!}{\includegraphics{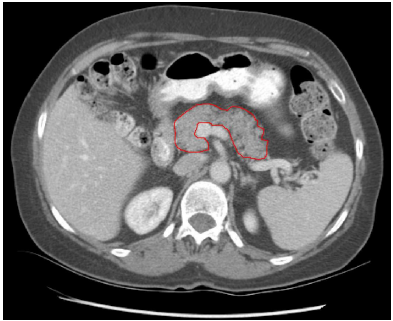}}
	\caption{Axial CT slice of a manual (gold standard) segmentation of the pancreas. The shape and size of the pancreas can vary drastically between patients. Furthermore, densities within the pancreas can vary and the contrast to surrounding tissues can be low in CT.}					
	\label{fig:pancreas_ct}
\end{figure}

Segmentation of the pancreas is an important input for many computer aided diagnosis (CADx) systems that could provide quantitative analysis, e.g. for diabetic patients. Accurate segmentation could also be necessary for other computer aided detection (CADe) methodologies that aim to detect pancreatic cancer. The literature is rich for the automatic segmentation of numerous organs in CT scans with sensitivities larger 90\%, especially for organs such as liver, heart or kidneys. However, high accuracy in the automatic segmentation of the pancreas is a challenging task. The pancreas' shape, size and location in the abdomen can vary drastically from patient to patient. Visceral fat tissue around the pancreas can cause large variations in contrast along its boundaries in CT. These factors make accurate and robust segmentation of the pancreas challenging. Figure \ref{fig:pancreas_ct} illustrates the noted challenges with a CT slice and ground-truth pancreas segmentation that was established manually by an experienced radiologist (gold standard). We aim to replicate these segmentations using computer vision and medical image computing techniques.
%%%%%%%%%%%%%%%%%%%%%%%%%%%%%%%%%%%%%%%%%%%%%%%%%%%%%%%%%%%%%
%%%%%%%%%%%%%%%%%%%%%%%%%%%%%%%%%%%%%%%%%%%%%%%%%%%%%%%%%%%%%
\section{Methods}
\begin{figure}[htb!]%[htb!]
\centering	\resizebox{0.8\textwidth}{!}{\includegraphics{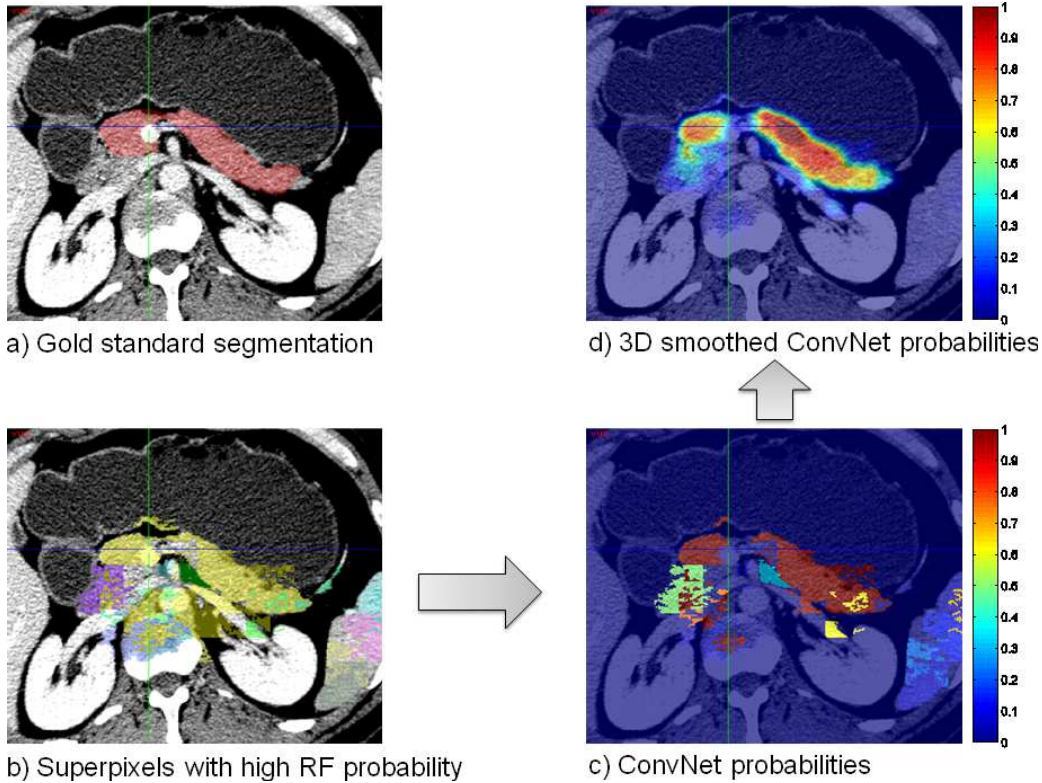}}
	\caption{Pancreas segmentation pipeline with (a) gold standard segmentation. Superpixels with high pancreas probability after random forest (RF) classification (b) are retained at cost of over-segmentation to serve as input to a convolutional network (ConvNet) classification (c). The ConvNet probabilities are then smoothed in 3D in order to obtain the final pancreas probability (d).}					
	\label{fig:pancreas_pipeline}
\end{figure}

Recently, the availability of large annotated training sets and the accessibility of affordable parallel computing resources via GPUs have made it feasible to train ``deep'' ConvNets (also popularized under the keyword: ``deep learning'') for computer vision classification tasks. ConvNets features are trained from the data in a fully supervised fashion. This has major advantages over more traditional CAD approaches that use hand-crafted features, designed from human experience. This means that ConvNets have a better chance of capturing the ``essence'' of the imaging data set used for training than when using hand-crafted features \cite{jones2014computer}. Great advances in classification of natural images have been achieved \cite{krizhevsky2012imagenet, zeiler2013visualizing}. Studies that have tried to apply deep learning and ConvNets to medical imaging applications also showed promise, e.g. \cite{Prasoon2013deep,roth2014new,roth2014detection}. In particular, ConvNets have been applied successfully in biomedical applications such as digital pathology \cite{cirecsan2013mitosis}. In this work, we apply ConvNets for pancreas segmentation. Our motivation is partially inspired by the spirit of hybrid systems using both parametric and non-parametric models for hierarchical coarse-to-fine classification using ConvNets \cite{girshick2013rich}. This hierarchical segmentation pipeline is illustrated in Fig. \ref{fig:pancreas_pipeline}.
%%%%%%%%%%%%%%%%%%%%%%%%%%%%%%%%%%%%%%%%%%%%%%%%%%%%%%%%%%%%%
\subsection{Superpixel candidate generation}
Our hierarchical method generates a set of local image regions (superpixels) $S = {S_1,…,S_N }$. All $N$ superpixels are extracted from the abdominal region using Simple Linear Iterative Clustering (SLIC) \cite{achanta2012slic}. Next, a 2D patch-level feature extraction and two-level cascade of random forest (RF) classifiers is implemented \cite{farag2014bottom}. Each patch is classified and probability response map is generated, where high probability patches reflect higher potential areas for pancreas tissue. The response maps are used to retain superpixel regions that consist of majority $p_\mathrm{RF} >0.5$. This results in a highly sensitive localization of superpixels $S_\mathrm{RF}$  within the pancreatic region but can cause vast over-segmentation. 
%%%%%%%%%%%%%%%%%%%%%%%%%%%%%%%%%%%%%%%%%%%%%%%%%%%%%%%%%%%%%
\subsection{Data augmentation}
\label{sec:augmentation}
Each retained superpixel region can serve as input to a ConvNet in order to assign a more distinct probability $p_\mathrm{ConvNet}$ of each superpixel in $S_\mathrm{RF}$ as being pancreas. Our ConvNet samples the bounding box of each superpixel $S_\mathrm{RF}^i$ at different scales $s$ and random non-rigid deformations $t$ (at training time). The degree of deformation is chosen such that the resulting warped images resemble plausible physical variations of the medical images. This approach is commonly referred to as data augmentation and can help avoid overfitting \cite{krizhevsky2012imagenet, cirecsan2013mitosis}.

A superpixel's bounding box is increased by a certain factor $s$ at each scale. Each non-rigid training deformation $t$ is computed by fitting a thin-plate-spline (TPS) to a regular grid of 2D control points $\left\{\omega_i;i=1,2,\ldots,K\right\}$. These control points can be randomly transformed at the 2D slice level and a deformed image can be generated using a radial basic function $\phi(r)$:
\begin{equation}
	t(x) = \sum^K_{i=1} c_i \phi\left(\left\|x-\omega_i\right\|\right).
\end{equation}
\begin{figure}[htb!]%[htb!]
\centering	\resizebox{0.9\textwidth}{!}{\includegraphics{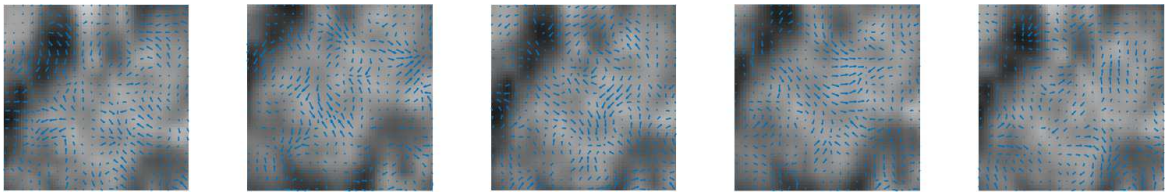}}
	\caption{We generate several random thin-plate-spline deformations in 2D in order to generate slight variations that are physically plausible in our training data. Some examples are shown here.}
	\label{fig:tps_deform}
\end{figure}
%%%%%%%%%%%%%%%%%%%%%%%%%%%%%%%%%%%%%%%%%%%%%%%%%%%%%%%%%%%%%
\subsection{Superpixel classification using ConvNets}
\begin{figure}[htb!]%[htb!]
\centering	\resizebox{0.9\textwidth}{!}{\includegraphics{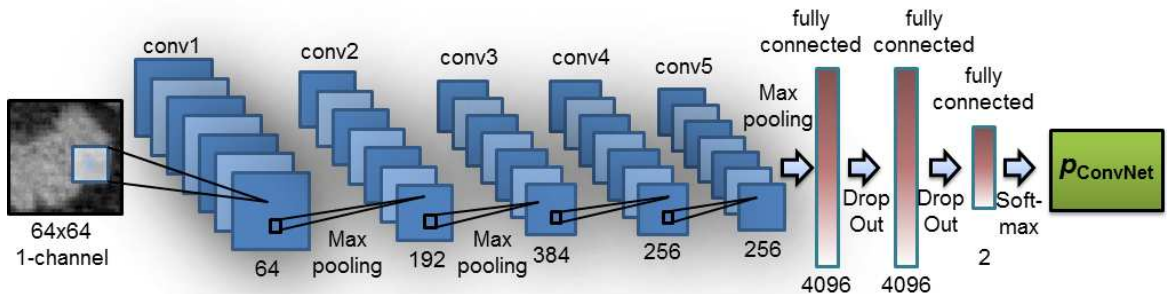}}
	\caption{The proposed ConvNet approach uses of five convolutional layers with max-pooling and locally fully-connected layers with DropOut connections. A final 2-way softmax layer is used for classification of pancreas and non-pancreas superpixels. The number of convolutional filters and neural network connections for each layer are as shown.}					
	\label{fig:convnet}
\end{figure}

\begin{figure}[htb!]%[htb!]
\centering	\resizebox{0.6\textwidth}{!}{\includegraphics{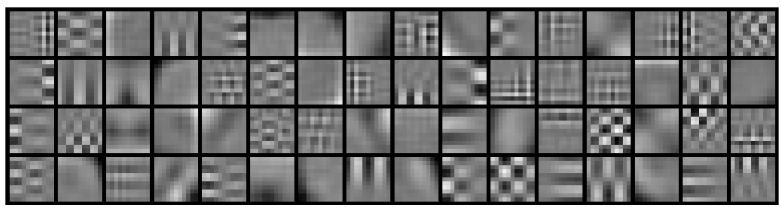}}
	\caption{The first layer of learned convolutional kernels of a ConvNet trained on superpixels extracted from CT images of the pancreas. Trained filters include simple shape enhancement filters and texture filters.}					
	\label{fig:conv1}
\end{figure}

The set of $N \times N_s \times N_t$ superpixel regions is used to train a ConvNet with a standard architecture for binary image classification. We use 5 cascaded layers of convolutional filters to compute image features. Other layers of the ConvNet perform max-pooling operations or consist of fully-connected neural networks. Our ConvNet ends with a final 2-way softmax layer for `pancreas' and `non-pancreas' classification (see Figure \ref{fig:convnet}). The fully connected layers are constrained in order to avoid overfitting. We use ``DropOut'' for this purpose. DropOut is a method that behaves as a regularizer when training the ConvNet \cite{hinton2012improving,srivastava2014dropout}. GPU acceleration allows efficient training of the ConvNet. We use an open-source implementation (\textit{cuda-convnet2}\footnote{\url{https://code.google.com/p/cuda-convnet2}}) by Krizhevsky et al. \cite{krizhevsky2012imagenet,krizhevsky2014one} which efficiently trains the ConvNet, using GPU acceleration. Further speed-ups are achieved using rectified linear units as neuron activation function instead of the traditional neuron model $f(x) = \tanh(x)$ or $f(x) = (1 + e^{-x})^{-1}$ in both training and testing \cite{krizhevsky2012imagenet}. 
The ConvNet automatically trains its convolutional filter kernels directly from the available training data. Examples of trained first-layer convolutional filters can be seen in Figure \ref{fig:conv1}. 
At testing, we evaluate each superpixel at $N_s$ different scales only in order to reduce redundancy and optimize computation time, rather than to also perform TPS deformations. This results in a probability for each superpixel being pancreas:
\begin{equation}
	p_\mathrm{ConvNet} = \left(x|p_1(x),\dots,p_{N_s}(x)\right) = \frac{1}{N_s }\sum_{i=1}^{N_s}p_i(x)
\end{equation}
The resulting per-superpixel ConvNet classifications $p_\mathrm{ConvNet}$ can then be assigned to each pixel in a superpixel region $S_\mathrm{RF}$ (see Figure \ref{fig:pancreas_pipeline}(b-c)), resulting in a probability map $P(x)$. Subsequently, we perform 3D filtering in order to average ConvNet probability across CT slices and neighboring regions, using a Gaussian kernel:
\begin{equation}
	G(x) =  \frac{1}{\sigma\sqrt{2\pi}} \exp{\left(\frac{-x^2}{2\sigma^2}\right)}
\end{equation}
Here, $\sigma$ defines the size of the Gaussian kernel, resulting in a probability map $G(P(x))$. This approach results in a smoother pancreas probability map as can be seen in Figure \ref{fig:pancreas_pipeline}(d). This step also propagates 2D probabilities to 3D by taking local 3D neighborhoods into account.
%%%%%%%%%%%%%%%%%%%%%%%%%%%%%%%%%%%%%%%%%%%%%%%%%%%%%%%%%%%%%
%%%%%%%%%%%%%%%%%%%%%%%%%%%%%%%%%%%%%%%%%%%%%%%%%%%%%%%%%%%%%
\section{Results}
Manual tracings of the pancreas for \Npatients post contrast abdominal CT volumes were provided by an experienced radiologist (gold standard segmentation). We use a random subset of \Ntrain cases to train a ConvNet in a supervised fashion and reserved \Nvalid cases for validation, and \Ntest cases for testing. We computed the optimally achievable superpixel classification based on the ground truth labels: average Dice of 80\% $\pm$ 4\% (range, 64-87\%) in training (Table \ref{tab:train_results}) and 81\% $\pm$ 3\% (range, 75-89\%) in testing (Table \ref{tab:test_results}). The optimal superpixel labeling is limited by the ability of superpixels to capture the true pancreas boundaries. This optimal labeling is used for assessing `positive' and `negative' superpixel examples for training. Furthermore, the training data is artificially increased by a factor $N_s \times N_t$ using the described data augmentation approach with both scale and random TPS deformations (see Sec. \ref{sec:augmentation}). Here, we train on an augmented data set using $N_s=2$, $N_t=8$. In testing we use $N_s=4$, $N_t=0$ and $\sigma=3$ voxels for computing smoothed probability maps $G(P(x))$. 

\begin{figure}[htb!]%[htb!]
\centering	\resizebox{0.5\textwidth}{!}{\includegraphics{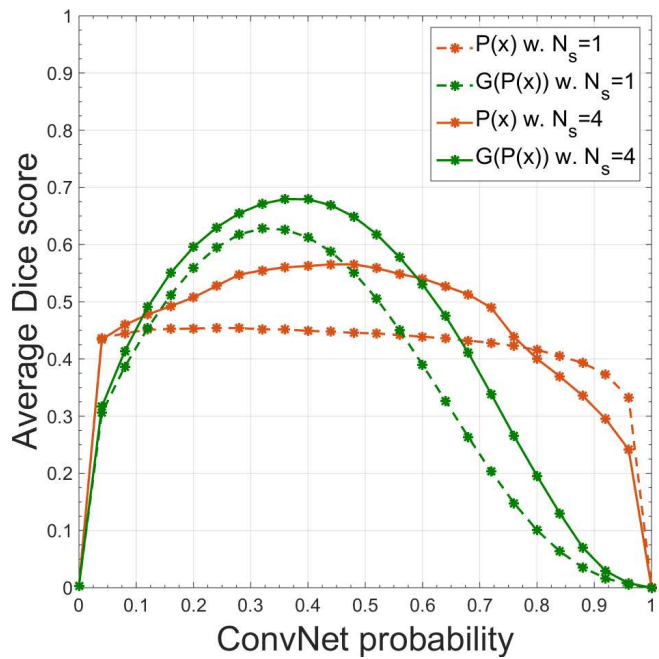}}
	\caption{Average Dice scores as a function of unsmoothed and 3D smoothed $p_\mathrm{ConvNet}$ probabilities at scales $N_s=1$ and $N_s=4$ in testing.}					
	\label{fig:froc}
\end{figure}

\begin{figure}[htb!]%[htb!]
\centering	\resizebox{0.9\textwidth}{!}{\includegraphics{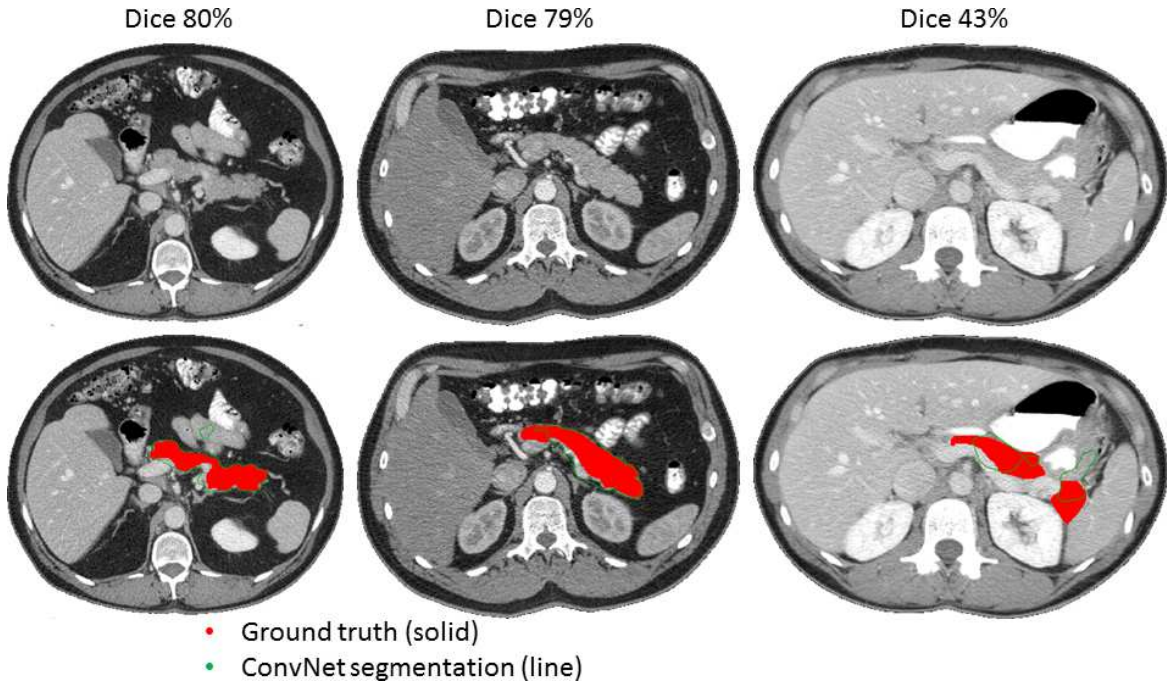}}
	\caption{Examples of pancreas segmentation using the proposed ConvNet approach (green outline). Red solid denotes manual ground truth annotations. The Dice scores are shown for two well segmented pancreases (left and middle) and one example where the segmentation leaked into neighboring organs (right). This poorer performance is likely associated to the lesser amount of visceral fat present in this patient, causing the boundaries between pancreas and surrounding tissues to be less well defined.}					
	\label{fig:axial_examples}
\end{figure}

The initial superpixel candidate labeling achieve Dice scores of only 27\% $\pm$ 6\% (range, 16-42\%) in testing but had a high sensitivity for labeling the pancreas (by applying an over-segmentation). Figure \ref{fig:froc} shows the average Dice scores after using the proposed ConvNet approach as an function of $P(x)$ and $G(P(x))$ at scales $N_s=1$ and $N_s=4$. Both more observational scales and Gaussian 3D smoothing improved the average Dice scores in testing. It can be observed that 3D smoothing has a larger contribution to segmentation performance than adding more scales. A maximum average Dice can be observed at $p_\mathrm{ConvNet}=0.4$ in our validation set ($n$=\Nvalid) after 3D Gaussian smoothing at $N_s=4$. This is the operation point we choose for testing. Utilizing ConvNets we achieve improved average Dice at this operation point of 68\% $\pm$ 10\% (range, 43-80\%) on the test set, an improvement of 41\% compared to the initial superpixel candidate labeling. This improvement is reflected in Table \ref{tab:test_results} that summarizes the Dice scores for each processing step of the algorithm. A marked improvement from 56\% to 68\% mean Dice score can be observed when applying the proposed 3D smoothing to ConvNet probabilities obtained from 2D superpixel classifications. We also show examples of axial pancreas segmentation based on the proposed ConvNet method at $p_\mathrm{ConvNet}=0.3$ in Fig. \ref{fig:axial_examples}. For comparison, the average Dice scores of the method on the training set are shown in Table \ref{tab:train_results}.

Training a ConvNet with $N \times N_s \times N_t=855,500$ example superpixel images of size $64\times 64$ pixels took 55 hours for 100 epochs on a modern GPU (NVIDIA GTX TITAN Z). However, execution time in testing is in the order of only 1 to 3 minutes per CT volume depending on the number of scales $N_s$.
%%%%%%%%%%%%%%%%%%%%%%%%%%%%%%%%%%%%%%%%%%%%%%%%%%%%%%%%%%%%%

\clearpage
\newpage
\begin{table}[htbp]
  \centering
  \caption{Training set: mean of optimally achievable Dice scores, our initial superpixel candidate labeling using $S_{RF}$, mean Dice scores on $P(x)$ and smoothed $G(P(x))$ using the proposed method with $N_s=4$ scales.}
    \begin{tabular}{lrrrr}
    \toprule
		\toprule
    \textbf{Patient}  & \textbf{Optimal} & Input $S_{RF}(x)$ & $P(x)$     & $G(P(x))$ \\
    \textbf{Training} & \textbf{}        &                   & w. $N_s=4$ & w. $N_s=4$ \\
		\midrule
    \textbf{Mean} & 0.80  & 0.26  & 0.69  & 0.79 \\
    \textbf{Std.} & 0.04  & 0.07  & 0.07  & 0.06 \\
    \textbf{Min.} & 0.64  & 0.14  & 0.35  & 0.39 \\
    \textbf{Max.} & 0.87  & 0.46  & 0.80  & 0.86 \\
    \bottomrule
		\bottomrule
    \end{tabular}%
  \label{tab:train_results}%
\end{table}%
%%%%%%%%%%%%%%%%%%%%%%%%%%%%%%%%%%%%%%%%%%%%%%%%%%%%%%%%%%%%%
\begin{table}[htbp]
 \small
  \centering
  \caption{Testing set: mean of optimally achievable Dice scores, our initial superpixel candidate labeling using $S_{RF}$, mean Dice scores on $P(x)$ and smoothed $G(P(x))$ using the proposed method at scales $N_s=1$ and $N_s=4$.}
    \begin{tabular}{lrrrrrr}
    \toprule
		\toprule
    \textbf{Patient} & \textbf{Optimal} & Input $S_{RF}$(x) & $P(x)$     & $G(P(x))$  & $P(x)$     & $G(P(x))$ \\
    \textbf{Testing} & \textbf{}        &                   & w. $N_s=1$ & w. $N_s=1$ & w. $N_s=4$ & w. $N_s=4$ \\
    \midrule
    \textbf{Mean} & 0.81  & 0.27  & 0.46  & 0.62  & 0.57  & 0.68 \\
    \textbf{Std.} & 0.03  & 0.06  & 0.10  & 0.14  & 0.09  & 0.10 \\
    \textbf{Min.} & 0.75  & 0.16  & 0.25  & 0.35  & 0.39  & 0.43 \\
    \textbf{Max.} & 0.89  & 0.42  & 0.58  & 0.76  & 0.67  & 0.80 \\
    \bottomrule
		\bottomrule
    \end{tabular}%
  \label{tab:test_results}%
\end{table}%
%%%%%%%%%%%%%%%%%%%%%%%%%%%%%%%%%%%%%%%%%%%%%%%%%%%%%%%%%%%%%
%%%%%%%%%%%%%%%%%%%%%%%%%%%%%%%%%%%%%%%%%%%%%%%%%%%%%%%%%%%%%
\clearpage
\newpage
\section{Conclusions}
This work demonstrates that ConvNets can be generalized to tasks in medical image analysis such as the segmentation of the pancreas. We show that superpixels can be classified into different tissue-types (pancreas and non-pancreas). Different scales and random non-rigid deformations of each superpixel region improve the ConvNet's classification performance. 

The results for the segmentation of the pancreas show promise, despite its simple nature of employing deep ConvNets for image superpixel classification. Different scales and Gaussian smoothing strategies are addressed and evaluated. We performed similar or better to the recent state-of-the art work of pancreas segmentation that reports average Dice scores ranging from 46.6\% to 68.8\% \cite{okada2012abdominal,wolz2013automated,farag2014bottom}. Other abdomen organs are jointly segmented in a multiple atlas fusion framework \cite{okada2012abdominal,wolz2013automated} to boost each other’s segmentation accuracy whereas we formulate the pancreas segmentation task as a standalone foreground/background separation problem. Note that \cite{okada2012abdominal, wolz2013automated} use leave-one-patient-out cross validation which is relatively computation expensive and may not scale up efficiently a large patient population. In particular, \cite{wolz2013automated} reports 68\% Dice coefficient with 150 patients and it drops significantly to 58\% when having only 50 patients under leave-one-out testing. Our results are based on a random \Ntrain/\Nvalid/\Ntest training and testing split. 

The proposed segmentation approach could be incorporated into a multi-organ segmentation method by specifying more tissue types as ConvNet supports multi-class classifications \cite{krizhevsky2012imagenet}. ConvNets could be especially useful for automatically learning to identify well-separable features for classifying multiple types of tissues using a similar approach as presented here.
%%%%%%%%%%%%%%%%%%%%%%%%%%%%%%%%%%%%%%%%%%%%%%%%%%%%%%%%%%%%%
%%%%%%%%%%%%%%%%%%%%%%%%%%%%%%%%%%%%%%%%%%%%%%%%%%%%%%%%%%%%%
\paragraph{Acknowledgments} This work was supported by the Intramural Research Program
of the NIH Clinical Center. This paper was presented at SPIE Medical Imaging 2015, Orlando, FL, USA. Refer to published version at \url{doi:10.1117/12.2081420}.
%%%%% References %%%%%
\bibliographystyle{spiebib}
\bibliography{references_spie2015}
\end{document}